\documentclass[letterpaper, 10 pt, conference]{ieeeconf}  

\IEEEoverridecommandlockouts                              

\usepackage{graphics} 
\usepackage{epsfig} 
\usepackage{mathptmx} 
\usepackage{times} 
\usepackage{amsmath} 
\usepackage{amssymb}  
\usepackage{xcolor}
\usepackage{hyperref}

\title{\LARGE \bf
Multi-Agent AI Framework for Road Situation Detection and C-ITS Message Generation}

\author{Kailin Tong$^{1}$, Selim Solmaz$^{1}$, Kenan Mujkic$^{1}$, Gottfried Allmer$^{2}$ and Bo Leng$^{3}$
	\thanks{$^{1}$ K. Tong, S. Solmaz and K. Mujkic are with Virtual Vehicle Research GmbH, Inffeldgasse 21a, 8010 Graz, Austria. {\tt\small \{kailin.tong, selim.solmaz, kenan.mujkic\}@v2c2.at.} 
	} 
    \thanks{$^{2}$ G. Allmer is with ASFINAG Maut Service GmbH,  Alpenstraße 99, 5020 Salzburg, Austria 	{\tt\small \{gottfried.allmer@asfinag.at\}.}}
    \thanks{$^{3}$ B. Leng is with Tongji University, 1239 Siping Road, Yangpu District, Shanghai, China. 	{\tt\small \{lengbo@tongji.edu.cn\}.}}
}

\begin{document}

\maketitle
\thispagestyle{empty}
\pagestyle{empty}

\begin{abstract}
Conventional road-situation detection methods achieve strong performance in predefined scenarios but fail in unseen cases and lack semantic interpretation, which is crucial for reliable traffic recommendations. This work introduces a multi-agent AI framework that combines multimodal large language models (MLLMs) with vision-based perception for road-situation monitoring. The framework processes camera feeds and coordinates dedicated agents for situation detection, distance estimation, decision-making, and Cooperative Intelligent Transport System (C-ITS) message generation. Evaluation is conducted on a custom dataset of 103 images extracted from 20 videos of the TAD dataset. Both Gemini-2.0-Flash and Gemini-2.5-Flash were evaluated. The results show $100\%$ recall in situation detection and perfect message schema correctness; however, both models suffer from false-positive detections and have reduced performance in terms of number of lanes, driving lane status and cause code. Surprisingly, Gemini-2.5-Flash, though more capable in general tasks, underperforms Gemini-2.0-Flash in detection accuracy and semantic understanding and incurs higher latency (Table II). These findings motivate further work on fine-tuning specialized LLMs or MLLMs tailored for intelligent transportation applications.

\end{abstract}

\section{Introduction}


Road safety remains a critical challenge in modern mobility. Despite advances in automated driving and Cooperative Intelligent Transport Systems (C-ITS), incidents such as accidents, adverse weather conditions, and hazardous road segments continue to cause significant societal and economic costs. While automated driving technologies promise safer and more efficient transportation, current systems often struggle to react promptly to diverse and unforeseen situations. This limitation underscores the importance of infrastructure-assisted monitoring and safety-warning mechanisms \cite{watzenig2016automated, Tong2020}.

Conventional detection approaches often employ specialized neural networks for accident recognition or anomaly detection. While effective in structured conditions, such methods fail to generalize to complex, open-world scenarios. Recent progress in multi-modal large language models (MLLMs) offers new opportunities for perception and reasoning in Intelligent Transportation Systems (ITS). With strong zero-shot and open-world detection capabilities, MLLMs are well suited for situation detection, where event types are unpredictable and responses must be situationally adaptive \cite{cui2024survey}. Yet, MLLMs alone lack domain-specific tuning and robust communication mechanisms.


The ESERCOM-D project addresses these challenges by introducing a multi-agent AI framework that combines MLLM-based perception and reasoning with infrastructure sensors for road situation detection and standardized C-ITS message generation. By coordinating agents for detection, distance estimation, decision-making, and Decentralized Environmental Notification Message (DENM) encoding, the framework enables infrastructure to act as an active safety system, enhancing responsiveness and resilience in complex traffic environments.

The main contributions of this paper is summarized as follows:

\begin{itemize}
    \item \textbf{Framework design:} We propose a multi-agent AI framework that integrates MLLMs with vision-based perception for road infrastructure monitoring. It generates standardized C-ITS messages in soft real time, with DENM schemas fully compliant with ETSI standards and directly integrable into existing systems.  
    \item \textbf{Empirical insights:} Using a real-world camera-image dataset, we benchmark two representative Google MLLMs on detection, information extraction, and decision-making. Results show that while the more advanced model offers broader general capabilities, its added complexity does not necessarily enhance structured ITS tasks and may instead increase latency.
    \item \textbf{Deployment perspective:} We discuss the practical strengths and limitations of the proposed framework, outlining a pathway toward real-world deployment.
\end{itemize}

The remainder of this paper is organized as follows. Section \ref{sec:related} reviews the related work. Section \ref{sec:problem} introduces the problem statement. Section \ref{sec:our} presents our approach in detail. Section \ref{sec:conclusion} reports and discusses the evaluation results. Finally, Section \ref{sec:conclusion} concludes the paper and provides an outlook on future research.

\section{Related Work} \label{sec:related}

Conventional vision-based models for traffic situation detection have achieved strong results in focused subdomains such as accident detection \cite{xu2024tad} and road-anomaly detection \cite{santhosh2020anomaly}. However, their performance often depends on domain-specific training data and they frequently lack robust generalization and explicit scene interpretation beyond closed-world settings. 

By contrast, MLLMs benefit from extensive pre-training and task-specific finetuning, enabling few- or even zero-shot learning, stronger out-of-distribution handling, and richer contextual reasoning over images and video \cite{yan2024forging}. This suggests a natural integration path into ITS for situation analysis and higher-level decision support. In parallel, Vehicle-to-Everything (V2X) communication provides road users and infrastructure with timely, standardized situational information about the environment and near-future routes, thereby augmenting perception and planning capabilities. Motivated by these trends, this work explores the joint use of MLLMs and V2X: we instantiate a practical pipeline for road-situation detection and the generation of standardized response messages for broadcast. In the following, we review the state of the art in both research directions.

\subsubsection{Traffic Situation Detection Using MLLMs}
Pioneering MLLM-driven efforts for driving scene understanding include DriveGPT4, which proposes an end-to-end autonomous driving and scene reasoning system using MLLM \cite{xu2024drivegpt4}.
Beyond scene description, MLLMs have been used for driver behavior analysis and risk assessment by integrating visual LLMs with reasoning chains \cite{zhang2024integrating}. Liao \emph{et~al.} introduced an MLLM-based grounding architecture that couples text, emotion, image, context, and cross-modal encoders with a multimodal decoder, demonstrating high prediction accuracy and operational efficiency on real-world benchmarks \cite{liao2024gpt}. Ilhan \emph{et~al.} examined prompt-enhancement strategies for improving interpretability and reported that certain forms of image “enhancement” can act as noise and fail to boost detection performance \cite{skender2025investigating}. 
Extending toward decision layers, MLLMs have also been leveraged to diagnose and repair issues in motion planning/trajectory generation pipelines for autonomous vehicles operating in dynamic environments \cite{lin2024drplanner}. Despite this progress, prior work has generally stopped at perception/understanding or diagnosis; relatively few studies couple direct traffic situation detection with standardized response-message generation suitable for V2X broadcast.

\subsubsection{Response-Message Generation}

An early pipeline proposed by Tong \emph{et~al.} that explicitly connects Connected and Automated Vehicles (CAVs) with large language models for incident detection/interpretation and C-ITS message generation is ConnectGPT \cite{tong2024connectgpt}. While the framework shows practical promise, broad validation at scale is limited by dataset size and by the rapid evolution of state-of-the-art MLLMs. Using a different response format, IncidentResponseGPT leverages generative AI to synthesize incident response plans and speed traffic-management recommendations; the authors also discuss risks such as data/guideline bias, ethics of autonomy, and the need for continuous validation \cite{grigorev2024incidentresponsegpt}. 
On the messaging infrastructure side, recent work from RWTH Aachen provides a software stack to use standardized ETSI ITS messages (e.g., CAM/ DENM) within ROS2-based systems and releases a public multi-modal V2X dataset (V2AIX), facilitating research that integrates perception/decision modules with communication stacks \cite{kueppers2024v2aix}. 

\section{Problem Statement} \label{sec:problem}

\begin{figure}
    \centering
    \includegraphics[width=1\columnwidth]{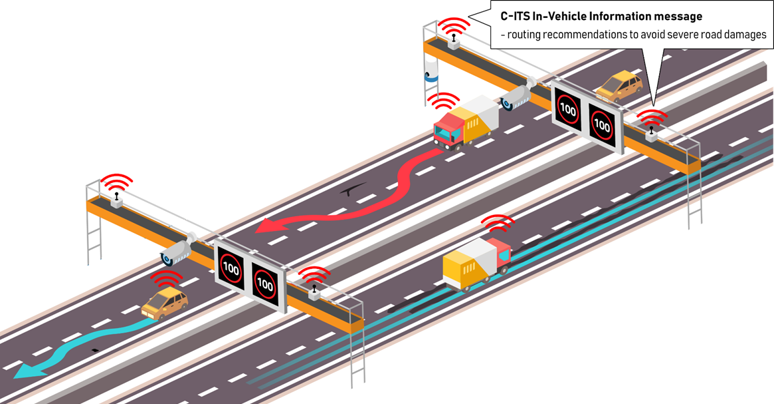}
    \caption{Exemplary scenarios: in-lane offset recommendation (cyan trajectory) and lane change recommendation (red trajectory) \cite{tong2024connectgpt}. }
    \label{fig:scenarios}
\end{figure}

Ensuring safe and efficient traffic flow remains a persistent challenge in the deployment of automated and connected vehicles. Prior efforts, such as the EU-H2020 ESRIUM project \cite{ESRIUMwebsite}, demonstrated the potential of infrastructure-assisted driving by creating precise digital maps of road surface deterioration and using them to guide lane positioning and routing strategies as illustrated in Fig. \ref{fig:scenarios}. While ESRIUM validated these concepts in large-scale on-road demonstrations\cite{Rudigier2022}, its scope was primarily restricted to road damage monitoring and did not address the dynamic management of unforeseen hazards or the generation of standardized C-ITS  messages.

In real-world traffic, imminent hazards—such as accidents, debris, adverse weather conditions, or stalled vehicles—require not only timely detection but also rapid and standardized communication to road users. Current approaches struggle to generalize beyond predefined hazard categories, often lacking mechanisms to translate perception into actionable, ETSI-compliant message formats. This gap highlights the need for frameworks that can operate robustly in open-world conditions while ensuring interoperability with existing traffic management and vehicle systems.

The ESERCOM-D project \cite{esercomd} directly addresses these challenges by developing a multi-agent AI framework that leverages MLLMs and infrastructure-based perception to automate the detection, interpretation, and communication of road situations. Our primary aim is to effectively alert oncoming CAVs (as well as connected vehicles in general) about impending dangers. Our methodology depends on infrastructure sensors, including cameras and potentially other perception sensors, in conjunction with a purpose-built MLLM serving as a central decision support tool. The central research question is therefore: How can we design an automated pipeline that ensures the reliable and timely generation of standardized C-ITS messages, providing oncoming CAVs with actionable warnings and guidance in unforeseen traffic scenarios?



\begin{figure*}
    \centering
    \includegraphics[width=2\columnwidth]{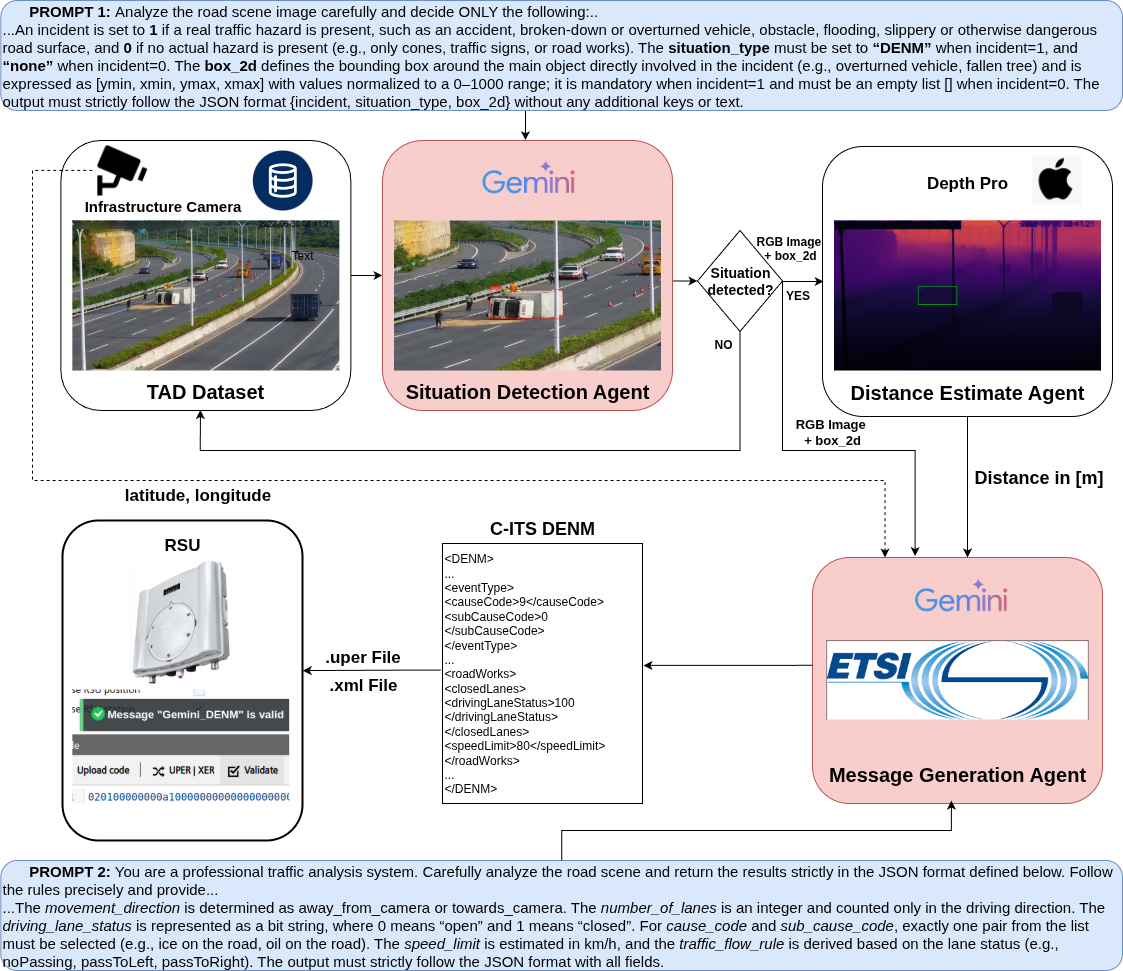}
    \caption{Pipeline of the proposed architecture: images from infrastructure cameras are analyzed by the Situation Detection Agent, distance is estimated by Depth Pro, and the Message Generation Agent produces standardized DENMs for transmission via Road Side Unit (RSU) devices.}
    \label{fig:pipeline}
\end{figure*}

\section{Our Approach} \label{sec:our}
Our architecture builds on autonomous agents that employ MLLMs to detect traffic situations and to automatically generate C-ITS messages in compliance with ETSI standards. The software backbone of the system is provided by LangChain \cite{chase2022langchain}, a framework designed to support the development of applications powered by LLMs.The entire process of the architecture is illustrated in Fig. \ref{fig:pipeline}. The first step is performed by the \textit{Situation Detection Agent}, which analyzes infrastructure camera images using predefined prompts. The agent determines whether an incident or hazard is present and assigns the corresponding \textit{situation\_type}. In case of a real event, the location of the object is additionally predicted through bounding box coordinates, which are then transformed and visually annotated on the image.\\
Once a situation is identified in the context, the \textit{Distance Estimate Agent} is triggered. It leverages Apple’s Depth Pro model \cite{apple2023depthpro} to estimate the distance in meters from the monocular camera input to the primary object associated with the situation. Subsequently, the \textit{Message Generation Agent} is activated and receives the estimated distance from the \textit{Distance Estimate Agent}. This agent employs the multimodal Gemini model \cite{google2023gemini} to extract additional parameters required for constructing a standardized C-ITS message, including the lateral and longitudinal location of the camera, the distance to the situation, the number and status of lanes, the cause and sub-cause codes, the estimated speed limit, and the traffic flow rule. Finally, the agent composes a structured object containing the \textit{management}, \textit{situation}, and \textit{alacarte} sections, resulting in a valid DENM.

To ensure interoperability, the \textit{Message Generation Agent} implements ASN.1 encoding using the \texttt{asn1tools} library, producing a UPER-encoded payload that is ready for transmission via RSU devices. The results are stored both as JSON and UPER text files, enabling both human and machine readability. The validity of the generated DENMs was further verified using the online ASN.1 decoder tool available at: \url{www.marben-products.com/decoder-asn1-automotive}.\\
In addition, the system integrates the \textit{LangSmith API key} \cite{langchain2023langsmith}, which enables advanced analysis and monitoring of requests, including latency, token usage, and overall agent reliability. This ensures improved performance evaluation.

\section{Evaluation} \label{sec:evaluation}

To conduct the evaluation, we extracted frames from all videos in the TAD benchmark dataset, which is based on China’s highway system \cite{xu2024tad}, covering both the training and test sets. Some videos of TAD dataset contain human faces, unrecognizable content, or scenes of complex highway structures. As these cases are unsuitable for applications in European traffic networks, they were filtered out. To further reduce redundancy, we excluded near-duplicate frames, since those within a 10-second video segment are largely indistinguishable. After manual filtering, we compiled a subset of 103 representative frames. This curated dataset served as the basis for evaluating Gemini-flash-2.0 and Gemini-flash-2.5 across the defined tasks.


\begin{table*}[htbp]  
\caption{Pipeline evaluation based on 103 images sampled from 20 different videos in the TAD dataset \cite{xu2024tad}.}
\label{tab:quantitative}
\begin{center}
\begin{tabular}{|l|c|c|c|c|c|c|c|c|}
\hline
Task & \multicolumn{4}{c|}{Situation Detection} & Number of Lanes & Driving Lane Status & Cause Code & Message Schema \\
\hline
Metric & Accuracy & Recall & Precision & F1-Score & Accuracy & Accuracy & Accuracy & Accuracy \\
\hline
Gemini-2.0-flash & \textbf{96.12\%} & \textbf{100\%} & \textbf{92.98\%} & \textbf{96.36\%} & 56,31\% & \textbf{47.57\%} & \textbf{77.67\%} & \textbf{100\%} \\
\hline
Gemini-2.5-flash & 90,29\% & \textbf{100\%} & 84.13\% & 91.38\% & \textbf{62.14}\% & 44.66\% & 70.87\% & \textbf{100\%} \\
\hline
\end{tabular}
\end{center}
\end{table*}

\begin{table*}[htbp]
\caption{Average request statistics for Gemini-2.0-Flash and Gemini-2.5-Flash, based on all requests collected from LangSmith. For each image, either one or two requests were issued depending on whether the initial MLLM output indicated situation\_type = DENM.}
\label{tab:avg_stats}
\begin{center}
\begin{tabular}{|l|c|c|}
\hline
Model & Avg. Tokens per Request & Avg. Latency per Request \\
\hline
Gemini-2.0-Flash & \textbf{2386} & \textbf{2.64s} \\
\hline
Gemini-2.5-Flash & 2503 & 12.29s \\
\hline
\end{tabular}
\end{center}
\end{table*}


Table~\ref{tab:quantitative} presents the pipeline evaluation of Gemini-flash-2.0 and Gemini-flash-2.5 on the curated dataset across the defined tasks. 
\textit{Situation Detection} denotes the binary classification of whether a situation is present in an image and is evaluated using recall and precision. 
\textit{Number of Lanes} specifies the number of lanes in the monitored highway segment, while \textit{Driving Lane Status} encodes lane closures in a binary system, where `0' indicates a closed lane and `1' indicates an open lane according to \cite{etsi-denm}. 
\textit{Cause Code} refers to the event cause classification according to \cite{etsi-denm}, for example, code ``90'' denoting ``hazardous location -- surface condition''. 
\textit{Number of Lanes}, \textit{Driving Lane Status}, and \textit{Cause Code} are all evaluated in terms of accuracy. 
Finally, \textit{Message Schema} assesses the correctness of the generated message structure with respect to the defined schema, in compliance with the ETSI ITS standard for DENM~\cite{etsi-denm}.

With regard to the metrics in Table~\ref{tab:quantitative}, situation detection is computed for each image, and both models achieved perfect recall (100\%), meaning that all situations present in the dataset were correctly identified. However, Gemini-2.0-Flash outperformed Gemini-2.5-Flash in terms of precision (92.98\% vs. 84.13\%), which also led to a higher F1-Score (96.36\% vs. 91.38\%) and overall accuracy (96.12\% vs. 90.29\%). Nonetheless, both models produced false positives, reflecting an imbalance between their perfect recall and lower precision. The remaining metrics—cause code, driving lane status, and number of lanes—are evaluated only if the Situation Detection Agent identifies that a situation occurs in the image according to the situation container of the DENM. In that case, the Message Generation Agent is triggered, retrieves the corresponding parameter values from the MLLM, and compares them against the reference GT table. For these more complex attributes, both models showed limited accuracy, with Gemini-2.0-Flash achieving 56.31\% for the number of lanes and 47.57\% for driving lane status, compared to 62.14\% and 44.66\% for Gemini-2.5-Flash. In the case of cause code classification, Gemini-2.0-Flash again performed slightly better (77.67\%) compared to Gemini-2.5-Flash (70.87\%). Finally, both models achieved a perfect score in message schema validation (100\%), confirming that the generated DENM outputs consistently conform to the standardized message structure.

Table II presents the average number of tokens per request and the average latency per request for both models. These values were computed using data collected through the LangSmith platform \cite{langchain2023langsmith}, which enables detailed monitoring and evaluation of all executed runs. Based on these measurements, aggregated averages were calculated to highlight the performance differences between the Gemini-2.0-Flash and Gemini-2.5-Flash models. Specifically, Gemini-2.0-Flash achieved an average of 2,386 tokens per request with a substantially lower latency of 2.64 seconds, whereas Gemini-2.5-Flash generated an average of 2,503 tokens per request with a significantly higher latency of 12.29 seconds. In summary, although Gemini-2.5-Flash is the more advanced model, Gemini-2.0-Flash consistently achieved higher accuracy, better overall performance in most situation detection and message content generation tasks, and substantially lower latency. Combined with its more favorable token cost, Gemini-2.0-Flash is the preferable choice for further development. One possible reason is that Gemini-2.5-Flash may have been optimized for broader multimodal reasoning and safer outputs, which can come at the expense of accuracy in structured tasks. Its greater computational complexity also contributes to the significantly higher latency observed.



\section{Conclusion and Outlook} \label{sec:conclusion}
This work presented a multi-agent AI framework that combines MLLMs with infrastructure-based perception to enhance road safety through standardized C-ITS message generation. By integrating situation detection, distance estimation, and DENM encoding into a coherent pipeline, the framework demonstrated its ability to detect hazards and broadcast actionable information in compliance with ETSI standards. Evaluation on a generated image dataset derived from the TAD dataset \cite{xu2024tad} confirmed the feasibility of the approach, with both Gemini models achieving $100\%$ recall in identifying situations and consistently producing valid DENMs. These findings establish a strong proof-of-concept for the role of multimodal AI in cooperative intelligent transport systems.

Nonetheless, the study also revealed limitations in extracting detailed parameters such as lane status, number of lanes, and cause codes, where accuracy remains modest. Moreover, the latency measurements of the two models indicate that a more advanced model with broader capabilities sacrifices computational efficiency without necessarily improving performance on structured tasks in our setting. These findings highlight the need for further fine-tuning of a specialized model tailored to our use cases.

Within the ESERCOM-D project, several concrete next steps will address these challenges. First, targeted fine-tuning of LLMs or MLLMs on traffic-specific datasets will be pursued to improve computational efficiency and domain specialization for deployment in operational road networks. Second, the pipeline will be validated in real-life motorway demonstration scenarios, where infrastructure-mounted cameras and RSUs will provide a testbed for end-to-end performance assessment. 
Finally, the integration of the pipeline into an automated driving demonstrator will enable the system to react directly to infrastructure recommendations, bridging the gap between perception, communication, and vehicle control.


\addtolength{\textheight}{-2cm}   



\section*{ACKNOWLEDGMENT}
The work was supported by the project ESERCOM-D. The Project ESERCOM-D is funded by the European Union under grant agreement No 101180176. Views and opinions expressed are however those of the author(s) only and do not necessarily reflect those of the European Union or European Union Agency for the Space Programme (EUSPA). Neither the European Union nor the granting authority can be held responsible for them.
The publication was written at Virtual Vehicle Research GmbH in Graz and partially funded within the COMET K2 Competence Centers for Excellent Technologies by the Austrian Federal Ministry for Innovation, Mobility and Infrastructure (BMIMI), Austrian Federal Ministry for Economy, Energy and Tourism (BMWET), the Province of Styria (Dept. 12) and the Styrian Business Promotion Agency (SFG). The Austrian Research Promotion Agency (FFG) has been authorised for the programme management.

\bibliographystyle{IEEEtran}

\bibliography{ref.bib}

\end{document}